\documentclass[10pt,twocolumn,letterpaper]{article}
\usepackage[pagenumbers]{wacv}      

\definecolor{wacvblue}{rgb}{0.21,0.49,0.74}
\usepackage[pagebackref,breaklinks,colorlinks,allcolors=wacvblue]{hyperref}
\usepackage{algorithm}
\usepackage{algpseudocode}
\usepackage{booktabs}

\def\wacvPaperID{3337} 
\def\confName{WACV}
\def\confYear{2027}

\title{Non-Linear Motion from Two Frames: Cubic Hermite Trajectories for Video Frame Interpolation}

\author{
Md Rifat Ur Rahman$^{1*}$,
Md Abtahi Majeed Chowdhury$^{1*}$,
Md Sakib Hossain Shovon$^{2*}$, \\
Jaesik Choi$^{2}$,
Mohammad Ali Moni$^{3}$ \\[0.5em]
$^{1}$Bangladesh University of Engineering and Technology, \;
$^{2}$KAIST, \;
$^{3}$Charles Sturt University \\
{\small $^{*}$Equal contribution}
}

\begin{document}
\maketitle
\begin{abstract}
Two-frame video frame interpolation commonly resamples endpoint optical flow by the query
time, which selects a constant-velocity straight-line path even though the same endpoint
displacement is compatible with many intermediate trajectories. We present
\textbf{HermiteVFI}, which replaces independent query-time motion prediction with one
explicit continuous trajectory per pixel. The measured endpoint displacement fixes the two
trajectory positions, while two latent endpoint derivatives learned from the frame pair
control the interior path. A cubic Hermite curve is the minimum-degree polynomial that can
independently satisfy these two position and two derivative boundary quantities. Its residual
form contains linear motion exactly, structurally preserves the measured displacement
endpoints for any network output, and has coefficients that are predicted once per frame
pair and evaluated in closed form at arbitrary $t$. We further show that one interior query
time gives a rank-deficient constraint on the two free trajectory coefficients, whereas any
two distinct interior times are full rank at the representation level, motivating multi-time
supervision. Appearance-derived directional cues help infer the latent derivatives that
endpoint displacement alone cannot identify. On X4K1000FPS, SNU-FILM, and
Vimeo90K-Septuplet, the largest gains occur for difficult non-uniform motion and high
interpolation ratios. The amortization applies to trajectory-coefficient prediction; downstream
reversal, refinement, and synthesis remain query dependent.
\end{abstract}
    
\section{Introduction}
\label{sec:intro}

Video frame interpolation (VFI) synthesizes one or more frames between two consecutive
keyframes, supporting slow-motion capture, frame-rate conversion, animation upsampling,
and video compression \cite{vfi-survey,superslomo}. The dominant family is flow based: an
optical-flow network estimates endpoint correspondence, that displacement is resampled to
the query time, and a synthesis network warps and blends the keyframes
\cite{superslomo,softsplat,rife,amt,ema-vfi}. Progress has come largely from better flow
estimators \cite{raft,biformer} and synthesis \cite{softsplat,vfiformer}, while temporal
resampling is often still close to linear.

Scaling endpoint flow by $t$ chooses one particular path: constant-velocity motion along a
straight segment. This can fail when objects accelerate, turn, reverse, or when the camera
starts and stops \cite{qvi,jnmr}. More fundamentally, endpoint displacement alone does not
identify the intermediate route: infinitely many paths can share the same start and end
positions. The challenge is therefore not merely to obtain a more accurate endpoint flow,
but to introduce a useful hypothesis class for the unobserved path between its endpoints.

Existing non-linear methods pay for this extra freedom in different ways. One group widens
the temporal window: quadratic methods use four frames \cite{qvi,eqvi}, later work
regresses multi-stage quadratic motion or selects among motion models over several flows
\cite{jnmr,dutta2022}, spline methods fit tracked trajectories \cite{ctvfi}, and event-based
methods use an additional sensor stream \cite{timelens,timelens++,timetracker}. A second
group keeps two frames but models the path implicitly, through a distilled acceleration
branch \cite{iqvfi}, a query-conditioned coordinate network \cite{gimm}, or a bidirectional
motion field \cite{bimvfi}. These approaches can represent non-linear motion, but the
intermediate positions are not expressed as one low-dimensional analytic curve whose
boundary displacement is satisfied by construction.

We target that gap. \textbf{HermiteVFI} formulates two-frame intermediate motion as a
single explicit continuous trajectory whose free degrees of freedom are two latent endpoint
derivatives. The measured flow fixes the endpoint positions; the learned derivatives bend
the interior path. We use ``endpoint velocity'' for the derivative of this interpolation
trajectory, not for a directly observed physical velocity. A cubic Hermite polynomial is a
natural and minimal carrier: per motion component, two endpoint positions and two endpoint
derivatives impose four independent boundary quantities, and therefore require polynomial
degree at least three. The resulting curve structurally satisfies the measured displacement
endpoints, contains the linear path as an exact zero-residual case, and is fixed after one
query-independent coefficient prediction.

The remaining degrees of freedom cannot be identified from endpoint displacement alone, so
we infer them from appearance-derived features and a parameter-free structure-tensor
directional descriptor. The parameterization also dictates the supervision. At a single
interior time, reconstruction exposes only one linear combination of the two trajectory
coefficients. We show that any two distinct interior times make this coefficient system full
rank at the representation level; in practice we supervise several times per pair to remain
robust to visibility, correspondence, and synthesis ambiguity. We train on X4K1000FPS
\cite{xvfi} and evaluate there and on SNU-FILM \cite{cain} and Vimeo90K-Septuplet
\cite{vimeo}.

Our contributions are:
\begin{itemize}
\item We formulate two-frame intermediate motion as an explicit cubic trajectory
      parameterized by latent endpoint derivatives. The motion representation exactly
      satisfies the measured endpoint displacements and contains linear motion as a nested
      zero-residual special case.
\item We characterize the supervision required by this parameterization: one interior query
      time is rank deficient for its two free coefficients, whereas any two distinct interior
      times are full rank at the representation level. This analysis motivates multi-time
      sequence supervision.
\item We predict the trajectory coefficients once per frame pair and evaluate arbitrary
      query times analytically, separating learned motion estimation from query-time
      trajectory evaluation. Existing evaluations show the largest benefits for fast,
      non-uniform motion and high interpolation ratios.
\end{itemize}

\section{Related Work}
\label{sec:related_works}

\paragraph{Flow-based frame interpolation.}
Within the flow-based pipeline, methods differ in how motion is resampled and how frames are
composited. SuperSloMo \cite{superslomo} uses a bilateral approximation of intermediate
flow, SoftSplat \cite{softsplat} resolves many-to-one collisions by softmax splatting, and
RIFE \cite{rife} predicts intermediate flow directly. Methods aimed at multi-frame output
also reduce repeated work: XVFI \cite{xvfi} predicts arbitrary-time flow from a
scale-invariant pyramid, while M2M \cite{m2m} estimates a reusable motion field and splats it
to multiple targets. M2M therefore shares the goal of amortizing motion estimation, but the
reused field corresponds to a linear path. Our learned coefficients instead define one
non-linear analytic trajectory; evaluating an additional trajectory time is closed form,
while downstream synthesis still runs per query.

\paragraph{Non-linear motion.}
QVI \cite{qvi} assumes constant acceleration over four frames, EQVI \cite{eqvi} rectifies
that fit by least squares, and later work regresses multi-stage quadratic motion or selects
among motion models \cite{jnmr,dutta2022}. Spline methods fit trajectories to tracked points
\cite{ctvfi}, and event-based methods integrate a second sensor stream
\cite{timelens,timelens++,timetracker}. Two-frame methods instead model non-linearity
implicitly: IQ-VFI \cite{iqvfi} uses a distilled acceleration prior, GIMM \cite{gimm} a
coordinate network conditioned on a motion latent, and BiM-VFI \cite{bimvfi} a bidirectional
motion field with a content-aware timestep. InterpAny-Clearer \cite{interpany} replaces the
scalar $t$ with a distance map. Our novelty is not the classical Hermite basis itself, but
its use as a structured two-frame motion parameterization: endpoint displacement fixes the
boundary positions, two learned latent derivatives control the interior, and one set of
query-independent coefficients determines the whole trajectory.

\paragraph{Directional cues in appearance.}
Motion blur can encode information about motion during exposure. Prior work reconstructs a
short video from a blurred image \cite{jin2018}, jointly deblurs and interpolates low-frame-
rate input \cite{bin,demfi}, or decodes temporal correlation from blur \cite{bit}. We do not
assume that a structure tensor uniquely measures motion blur. Instead, we use local
anisotropy and dominant orientation as a parameter-free directional appearance cue that can
be informative under blur but can also respond to ordinary oriented image structure.

\paragraph{Supervision for non-linear motion.}
Prior two-frame methods often use privileged supervision: RIFE \cite{rife} distills
intermediate flow from a teacher that sees the ground-truth frame, IQ-VFI \cite{iqvfi} adds
acceleration and motion distillation, and BiM-VFI \cite{bimvfi} uses a knowledge-distillation
term. IQ-VFI also reports no gain from an acceleration branch trained by reconstruction
alone on Vimeo triplets. Its parameterization has one free coefficient per pixel, so the
specific two-coefficient rank deficiency analyzed in Sec.~\ref{sec:trajectory} does not
apply. More generally, a single supervised time does not test whether a learned correction
remains coherent at other times; our explicit trajectory makes this accountability visible
and motivates multi-time supervision.

\begin{figure*}[t]
\centering
\includegraphics[width=\textwidth]{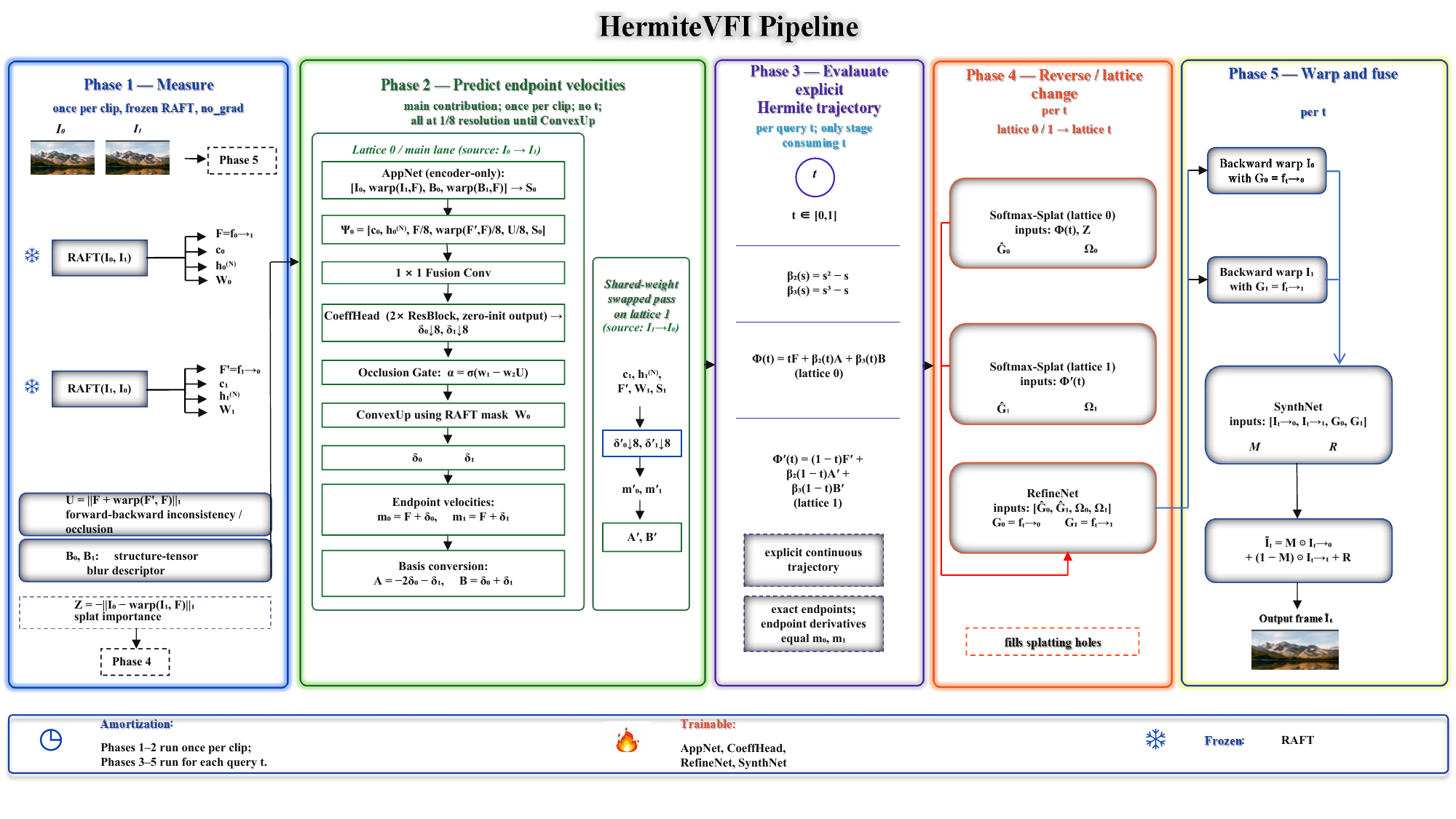}
\caption{\textbf{Overview of HermiteVFI.} A frozen flow estimator measures bidirectional
flow and auxiliary maps from the two keyframes. A shared head then predicts a velocity
residual at each keyframe, which fixes the coefficients of a cubic Hermite curve. The
curve is evaluated in closed form at each query time, splatted onto the target grid,
refined, and rendered by warping and blending. The head does not take $t$ as input and
runs once per frame pair; only the last three stages repeat.}
\label{fig:pipeline}
\end{figure*}

\section{Method}
\label{method}

Given adjacent frames $I_0, I_1 \in \mathbb{R}^{H\times W\times 3}$, flow-based VFI
estimates the motion between them, resamples it to a query time $t \in (0,1)$, and
synthesizes $\hat I_t$ by warping and blending. We focus on the resampling step. Let
$F = f_{0\to1}$ and $F' = f_{1\to0}$ be bidirectional flows from a pre-trained estimator,
$\Phi(t)$ the displacement of a pixel from its position in $I_0$ at time $t$, and
$\overleftarrow{\omega}, \overrightarrow{\omega}$ the backward warping and forward
splatting operators. Primes denote quantities defined on the pixel grid of $I_1$. Linear
motion models take $\Phi(t) = tF$.

\subsection{Hermite motion representation}
\label{sec:trajectory}

\paragraph{Latent endpoint derivatives and minimum degree.}
$F$ is the net displacement over $[0,1]$, not an observation of the instantaneous motion at
either endpoint. We parameterize the interpolation trajectory by its endpoint derivatives
$m_0 = \dot\Phi(0)$ and $m_1 = \dot\Phi(1)$. We call these quantities endpoint velocities
because they are derivatives of the learned trajectory; they are latent reconstruction-optimal
parameters rather than directly measured physical velocities. Together with
$\Phi(0)=\mathbf{0}$ and $\Phi(1)=F$, they specify four boundary quantities per motion
component. A degree-$d$ polynomial has $d+1$ coefficients, so satisfying four independent
position/derivative constraints requires $d\geq3$. The cubic Hermite curve is therefore the
minimum-degree polynomial supporting this formulation,
\begin{equation}
\Phi(t) = (t^3\!-\!2t^2\!+\!t)\,m_0 + (-2t^3\!+\!3t^2)\,F + (t^3\!-\!t^2)\,m_1 .
\label{eq:hermite}
\end{equation}

\paragraph{Residual form.}
We use a residual parameterization around the measured endpoint displacement,
$\delta_i=m_i-F$, so the network predicts only departure from the linear reference. Substituting into Eq.~\eqref{eq:hermite} and
collecting terms gives
\begin{equation}
\Phi(t) = t\,F + \beta_2(t)\,A + \beta_3(t)\,B,
\label{eq:traj}
\end{equation}
with $\beta_2(s) = s^2\!-\!s$, $\beta_3(s) = s^3\!-\!s$, and
\begin{equation}
A = -2\delta_0 - \delta_1, \qquad B = \delta_0 + \delta_1 .
\label{eq:coeffs}
\end{equation}
Eq.~\eqref{eq:traj} is an exact rewriting of Eq.~\eqref{eq:hermite}: the linear model is
the leading term and the network predicts only the deviation from it. Evaluating a
timestep costs three tensor products and no network pass. \emph{The full expansion is
given in Sec.~A.1 of the supplementary.}

\paragraph{Proposition 1: endpoint consistency and nested linear motion.}
Both basis functions vanish at $s=0$ and $s=1$, hence $\Phi(0)=\mathbf{0}$ and
$\Phi(1)=F$ for arbitrary network outputs $A,B$. This is an exact guarantee for the
displacement trajectory; it does not imply exact RGB reconstruction by the downstream
synthesis network. Differentiation gives $\dot\Phi(0)=F-A-B=m_0$ and
$\dot\Phi(1)=F+A+2B=m_1$. Setting $\delta_0=\delta_1=0$ yields $A=B=0$ and recovers
$\Phi(t)=tF$ exactly. Setting $\delta_1=-\delta_0$ instead eliminates $B$ and yields a
quadratic correction to the linear path, showing that lower-order motion is nested within
the same representation.

\paragraph{Bounded deviation.}
Write $\bar\delta = \tfrac12(\delta_0+\delta_1)$ for the common endpoint-derivative
component and $\Delta = \tfrac12(\delta_1-\delta_0)$ for their difference. Substituting
into Eq.~\eqref{eq:coeffs} gives $A=-3\bar\delta+\Delta$ and $B=2\bar\delta$, so
\begin{equation}
\Phi(t)-tF = \bar\delta\,t(2t\!-\!1)(t\!-\!1)+\Delta\,(t^2\!-\!t).
\label{eq:bound}
\end{equation}
\paragraph{Proposition 2: bounded departure from linear motion.}
For $g_c(t)=t(2t-1)(t-1)$ and $g_d(t)=t^2-t$,
$\max_{[0,1]}|g_c(t)|=\sqrt{3}/18\approx0.0962$ and
$\max_{[0,1]}|g_d(t)|=1/4$. Hence, for any norm obeying the triangle inequality,
\begin{equation}
\sup_{t\in[0,1]}\|\Phi(t)-tF\|
\leq \frac{\sqrt{3}}{18}\|\bar\delta\|+\frac14\|\Delta\|.
\label{eq:supbound}
\end{equation}
The differential component $\Delta$ therefore has approximately $2.6\times$ greater
worst-case basis amplitude than the common component $\bar\delta$; the latter also
vanishes at $t=1/2$. Endpoint-derivative disagreement has stronger leverage on path
bending, without being the only source of deviation. We report
$\rho(\delta_0,\delta_1)$ alongside $\|\delta_i\|$ in Sec.~\ref{exp}. Residual magnitudes
are measured in full-resolution pixels after upsampling.

\paragraph{Backward trajectory.}
The same construction on the grid of $I_1$ with time reversed gives
\begin{equation}
\Phi'(t) = (1\!-\!t)F' + \beta_2(1\!-\!t)A' + \beta_3(1\!-\!t)B' ,
\label{eq:traj-prime}
\end{equation}
with $A', B'$ formed from $\delta'_0, \delta'_1$ as in Eq.~\eqref{eq:coeffs}.

\paragraph{Proposition 3: representation-level identifiability.}
At one interior time $t$, Eq.~\eqref{eq:traj} exposes only one linear combination of $A$
and $B$, so the two coefficients are not uniquely identifiable. For two distinct interior
times $t_1,t_2$,
\begin{equation}
\begin{bmatrix}
\Phi(t_1)-t_1F\\[1pt]
\Phi(t_2)-t_2F
\end{bmatrix}
=
\begin{bmatrix}
\beta_2(t_1) & \beta_3(t_1)\\
\beta_2(t_2) & \beta_3(t_2)
\end{bmatrix}
\begin{bmatrix}A\\B\end{bmatrix}.
\label{eq:ident}
\end{equation}
Because $\beta_3(t)=\beta_2(t)(t+1)$, the determinant is
$\beta_2(t_1)\beta_2(t_2)(t_2-t_1)$, which is nonzero for
$t_1,t_2\in(0,1)$ with $t_1\neq t_2$. Thus any two distinct interior trajectory
observations uniquely determine $A,B$ at the representation level. Image reconstruction is
less direct because visibility, correspondence, occlusion, and synthesis introduce
additional ambiguity. The consistency term of Sec.~\ref{sec:loss} couples the two lattices
but is also satisfied by $\delta_i\equiv0$; in practice we therefore supervise several
interior times per pair. X4K1000FPS~\cite{xvfi} and Vimeo90K-Septuplet~\cite{vimeo}
provide the required multi-time sequences.

\subsection{Endpoint velocity estimation}
\label{sec:velocity}

\paragraph{Inputs.}
We run a frozen RAFT~\cite{raft} in both directions, retaining its context features $c_i$,
final hidden state $h_i$, and convex upsampling mask $W_i$ in addition to the flows. From
the flows we compute a forward--backward inconsistency map and a photometric residual,
\begin{equation}
U = \| F + \overleftarrow{\omega}(F', F) \|_1, \qquad
Z = -\| I_0 - \overleftarrow{\omega}(I_1, F) \|_1 ,
\label{eq:maps}
\end{equation}
used as an occlusion proxy and as splatting importance respectively.

\paragraph{Directional appearance descriptor.}
Endpoint displacement alone is insufficient to identify the intermediate route. We therefore
provide the velocity head with a parameter-free local directional cue. From the smoothed
structure tensor
$\mathcal{G}_{5\times5} * (\nabla I_i \nabla I_i^{\!\top})$, with eigenvalues
$\lambda_1 \ge \lambda_2$ and dominant orientation $\theta$, we form a parameter-free
descriptor
\begin{equation}
\mathcal{B}_i = \big[\, \lambda_1/(\lambda_2\!+\!\epsilon),\; \cos 2\theta,\; \sin 2\theta \,\big].
\label{eq:blur}
\end{equation}
The descriptor captures local anisotropy and dominant orientation. It can be informative in
regions affected by directional motion blur, but is not specific to blur and may also respond
to ordinary oriented image structure. \emph{The closed form of the eigenvalues and
double-angle terms is given in Sec.~A.3 of the supplementary.}

\paragraph{Velocity head.}
An appearance encoder extracts features from the frame pair and directional descriptors, warped
onto the grid of $I_0$,
\begin{equation}
\mathcal{S}_0 = \mathrm{AppNet}\big([\, I_0,\; \overleftarrow{\omega}(I_1, F),\;
\mathcal{B}_0,\; \overleftarrow{\omega}(\mathcal{B}_1, F) \,]\big),
\label{eq:appnet}
\end{equation}
which are concatenated with the flow-derived inputs at $1/8$ resolution,
\begin{equation}
\Psi_0 = \big[\, c_0,\; h_0,\; \tfrac18 F^{\downarrow 8},\;
\tfrac18 \overleftarrow{\omega}(F',F)^{\downarrow 8},\; U^{\downarrow 8},\;
\mathcal{S}_0 \,\big],
\label{eq:concat}
\end{equation}
and passed to a single-scale convolutional head $\mathcal{C}$ predicting both residuals,
\begin{equation}
\big[\delta_0^{\downarrow 8}, \delta_1^{\downarrow 8}\big] = \mathcal{C}(\Psi_0).
\label{eq:head}
\end{equation}
The output layer of $\mathcal{C}$ is zero-initialized, so training starts from the linear
model. Velocity is ill-defined under occlusion, so the residuals are gated at the
resolution they are predicted at, before upsampling,
\begin{equation}
\alpha = \sigma\big(w_1 - w_2 U^{\downarrow 8}\big), \qquad
\delta_i^{\downarrow 8} \leftarrow \alpha \odot \delta_i^{\downarrow 8} ,
\label{eq:gate}
\end{equation}
after which the gated residuals are carried to full resolution and added to the measured
flow,
\begin{equation}
\delta_i = \mathrm{ConvexUp}\big(8\,\delta_i^{\downarrow 8},\, \mathrm{softmax}(W_0)\big),
\qquad m_i = F + \delta_i .
\label{eq:upsample}
\end{equation}
Reusing $W_0$ keeps the residual aligned to the motion boundaries of $F$ at no parameter
cost, and gating first means the gate inherits those boundaries too rather than cutting
across them. Note that $\alpha$ is a two-parameter function of $U$ and can be evaluated at
whichever resolution a term requires. From $\delta_0, \delta_1$, $A$ and $B$ follow from
Eq.~\eqref{eq:coeffs}. The head is shared and applied a
second time to $\Psi_1$, built from the already-available backward-pass state, giving
$A', B'$. Neither pass depends on $t$, so both are evaluated once per frame pair.
\emph{Convex upsampling weights and layer specifications are given in Sec.~A.4 and
Sec.~B of the supplementary.}

\subsection{Integration with frame interpolation}
\label{sec:synthesis}

Eq.~\eqref{eq:traj} produces flow on the grid of $I_0$, so we transfer it to the target
grid by softmax splatting~\cite{softsplat} with negated displacement,
\begin{equation}
\tilde G_0 = \overrightarrow{\omega}\big(-\Phi(t),\, \Phi(t),\, Z\big), \qquad
\tilde G_1 = \overrightarrow{\omega}\big(-\Phi'(t),\, \Phi'(t),\, Z'\big),
\label{eq:splat}
\end{equation}
and fill unsplatted pixels, identified by their accumulated weight $\Omega_i$, with a
refinement network,
\begin{equation}
G_i = \tilde G_i + \mathrm{RefineNet}(\tilde G_0, \tilde G_1, \Omega_0, \Omega_1).
\label{eq:refine}
\end{equation}
\emph{The per-pixel form of Eq.~\eqref{eq:splat} is given in Sec.~A.5 of the
supplementary.} Following flow-based VFI practice~\cite{softsplat,amt,gimm}, the frames
are backward warped, $I_{t\to i} = \overleftarrow{\omega}(I_i, G_i)$, and fused by a mask
and residual,
\begin{equation}
M, R = \mathrm{SynthNet}(I_{t\to0}, I_{t\to1}, G_0, G_1),
\label{eq:synth}
\end{equation}
\begin{equation}
\hat I_t = M \odot I_{t\to0} + (1\!-\!M) \odot I_{t\to1} + R .
\label{eq:fuse}
\end{equation}
\emph{Pseudocode for inference and for a training iteration is given in Sec.~C of the
supplementary.}

\subsection{Loss}
\label{sec:loss}

We supervise a set $\mathcal{T}$ of query times per frame pair, sharing one coefficient
prediction across all of them. The velocity-consistency term is normalized by the per-sample
motion scale $s=\max(\|F\|_\infty,\|F'\|_\infty)$ so that its weight does not grow
with clip speed.

\paragraph{Reconstruction.}
Predictions are scored by a photometric bundle $\mathcal{L}_{\text{photo}}$ combining a
Charbonnier penalty, a Laplacian pyramid loss, the census loss~\cite{rife}, and
LPIPS~\cite{lpips} at equal weight. We apply it to $\hat I_t$ and, at half weight, to the
residual-free blend $\tilde I_t = M \odot I_{t\to0} + (1\!-\!M)\odot I_{t\to1}$,
\begin{equation}
\mathcal{L}_{\text{rec}} = \frac{1}{|\mathcal{T}|} \sum_{t\in\mathcal{T}}
\Big[ \mathcal{L}_{\text{photo}}(\hat I_t, I_t)
+ \tfrac12 \mathcal{L}_{\text{photo}}(\tilde I_t, I_t) \Big].
\label{eq:rec}
\end{equation}
The auxiliary term prevents the synthesis residual $R$ from repainting the frame unaided,
which would starve the motion path of gradient. \emph{The bundle terms are defined in
Sec.~A.6 of the supplementary.}

\paragraph{Velocity consistency.}
Eq.~\eqref{eq:traj} and Eq.~\eqref{eq:traj-prime} describe one motion from opposite ends,
so a pixel leaving $\mathbf{x}$ and arriving at $\mathbf{x} + F(\mathbf{x})$ requires
$m_1 = -\overleftarrow{\omega}(m'_0, F)$ and $m_0 = -\overleftarrow{\omega}(m'_1, F')$. No
correspondence exists where a pixel is occluded, so each residual is weighted by the
reliability map of the lattice it lives on, $\alpha$ for the first and $\alpha'$ for the
second,
\begin{equation}
\begin{split}
\mathcal{L}_{\text{vel}} = \;
& \tfrac12 \big\langle \alpha \odot
\big| m_1 + \overleftarrow{\omega}(m'_0, F) \big| / s \big\rangle \\
+ \; & \tfrac12 \big\langle \alpha' \odot
\big| m_0 + \overleftarrow{\omega}(m'_1, F') \big| / s \big\rangle .
\end{split}
\label{eq:vel}
\end{equation}
with $\langle\cdot\rangle$ averaging over pixels and batch. This is the same $\alpha$ of
Eq.~\eqref{eq:gate}: it is small where the forward--backward check disagrees, so the
constraint is applied where a correspondence exists and released where it does not. This is a consistency
constraint rather than a source of curvature, as $\delta_i \equiv 0$ satisfies it exactly.
\emph{The derivation is given in Sec.~A.2 of the supplementary.} The full objective is
$\mathcal{L} = \mathcal{L}_{\text{rec}} + \lambda_{\text{vel}} \mathcal{L}_{\text{vel}}$.

\begin{table*}[htbp]
\centering
\caption{Intermediate motion, PSNR$_f\uparrow$ / EPE$\downarrow$. PSNR$_f$ is computed on
normalized flows and EPE on unscaled flows. \#F is the number of input frames. Ours-V is
trained on Vimeo90K-Septuplet and Ours-X on X4K1000FPS; we evaluate both on all
benchmarks.}
\label{tab:flow}
\begin{tabular*}{\textwidth}{@{\extracolsep{\fill}}lcccc@{}}
\toprule
& & \multicolumn{2}{c}{X-TEST} & VSF \\
\cmidrule(lr){3-4}\cmidrule(lr){5-5}
Method & \#F & $8\times$ & $16\times$ & \\
\midrule
Linear \cite{superslomo}     & 2        & 29.43 / 1.88 & 25.04 / 3.49 & 30.06 / 2.89 \\
QVI \cite{qvi}               & 4        & 32.35 / 1.16 & 28.10 / 2.23 & 30.10 / 2.84 \\
EQVI \cite{eqvi}             & 4        & 32.76 / 1.11 & 28.43 / 2.10 & 30.15 / 2.81 \\
IQ-VFI \cite{iqvfi}          & 2        & 33.49 / 0.98 & 29.14 / 1.89 & 30.24 / 2.75 \\
GIMM \cite{gimm}             & 2        & 37.05 / 0.38 & 33.09 / 1.11 & 30.45 / 2.68 \\
CT-VFI \cite{ctvfi}          & 2$^\ast$ & 29.52 / 1.85 & 25.11 / 3.46 & 30.09 / 2.88 \\
CT-VFI (spline) \cite{ctvfi} & $N$      & 37.26 / 0.35 & 33.45 / 1.06 & 30.48 / 2.65 \\
\midrule
Ours-V (linear)              & 2        & 29.46 / 1.87 & 25.07 / 3.48 & 30.07 / 2.88 \\
Ours-V                       & 2        & 36.84 / 0.42 & 32.90 / 1.14 & \textbf{30.64} / \textbf{2.56} \\
Ours-X (linear)              & 2        & 29.46 / 1.87 & 25.07 / 3.48 & 30.07 / 2.88 \\
Ours-X                       & 2        & \textbf{37.78} / \textbf{0.32} & \textbf{34.19} / \textbf{0.97} & 30.43 / 2.69 \\
\bottomrule
\end{tabular*}
\end{table*}

\section{Experiments}
\label{exp}

\subsection{Setup}

We train two models. Ours-V is trained on Vimeo90K-Septuplet \cite{vimeo}, where motion is
slow and close to linear. Ours-X is trained on X4K1000FPS \cite{xvfi}, where it is fast and
non-uniform. Each model is trained once and run on every benchmark, so both are tested in
and out of distribution. Ours (linear) fixes $\delta_i \equiv 0$ and is otherwise identical.

We evaluate motion and frame quality separately. For motion we report PSNR$_f$ on
normalized flows and EPE on unscaled flows, following \cite{gimm}. On Vimeo-Septuplet-Flow
the pseudo-ground-truth comes from FlowFormer \cite{flowformer}. On X-TEST-Flow we instead
chain small-displacement flows through the native-rate frames, which is more reliable than
one large-displacement estimate under this benchmark's fast motion. For frames we report
PSNR and LPIPS \cite{lpips} on X-TEST, SNU-FILM\cite{cain}, and Vimeo90K-Septuplet.

The baselines are RIFE \cite{rife} and EMA-VFI \cite{ema-vfi} for linear motion, QVI
\cite{qvi} and EQVI \cite{eqvi} for four-frame quadratic motion, IQ-VFI \cite{iqvfi} and
GIMM-VFI \cite{gimm} for two-frame non-linear motion, and CT-VFI \cite{ctvfi} in both its
two-frame tracked setting ($2^\ast$) and its full-sequence spline setting ($N$). We run all
of them with their released weights. The \#F column gives the number of input frames each
method needs.

GIMM-VFI is the closest external comparison because it also models non-linear motion from
two frames, but it is not architecture or training matched: its synthesis module derives
from AMT-G, ours from splatting-based synthesis~\cite{softsplat}, and its released weights
are trained on Vimeo90K-Septuplet. Our photometric bundle also adds LPIPS to the
Charbonnier--Laplacian--census objective of~\cite{gimm}; because LPIPS is an evaluation
metric, we additionally report \emph{Ours-X (no LPIPS)}. Table~\ref{tab:flow} evaluates
intermediate motion before our synthesis stage and therefore reduces one source of
system-level confounding, but it is not a controlled estimate of the Hermite
parameterization alone. The matched linear-versus-Hermite variants within our own pipeline
provide the cleanest controlled evidence for the trajectory representation; external rows
measure complete methods under their released settings.

Training uses AdamW, learning rate $2\times10^{-4}$, batch size 32, for 60 epochs on $2$ GPUs at $256\times256$ crops. All supervised $t$ within a pair share one trajectory-coefficient prediction.

\textit{Additional qualitative comparisons are provided in sec. E of the supplementary.}

\subsection{Intermediate motion}

Table~\ref{tab:flow} evaluates intermediate motion before synthesis. Within the matched
pipeline, replacing the fixed linear path by the learned Hermite trajectory reduces X-TEST
$16\times$ EPE from 3.48 to 0.97 for Ours-X. Under their released settings, the external
comparisons are 1.11 for GIMM and 2.23/2.10 for the four-frame quadratic methods. CT-VFI's
full-sequence cubic-spline setting reaches 1.06 but requires the complete sequence, whereas
Ours-X uses two frames. On VSF the margin over our linear control is small, consistent with
the near-linear motion of that benchmark. We interpret the within-system linear-to-Hermite
gap causally; differences to external methods remain system-level comparisons.

The two linear rows are identical because they are the same trajectory. Fixing
$\delta_i \equiv 0$ leaves no trained parameter in $\Phi$, so $\Phi(t) = tF$ comes entirely
from the frozen estimator and cannot depend on which dataset trained the head. We list them
separately only to pair each with its non-linear counterpart.

\begin{table*}[htbp]
\centering
\small
\caption{Frame interpolation, PSNR$\uparrow$ / LPIPS$\downarrow$. Notation as in
Table~\ref{tab:flow}.}
\label{tab:main}
\begin{tabular*}{\textwidth}{@{\extracolsep{\fill}}lccccccc@{}}
\toprule
& & \multicolumn{2}{c}{X-TEST} & \multicolumn{3}{c}{SNU-FILM-arb} & VSF \\
\cmidrule(lr){3-4}\cmidrule(lr){5-7}\cmidrule(lr){8-8}
Method & \#F & $8\times$ & $16\times$ & Medium ($4\times$) & Hard ($8\times$) & Extreme ($16\times$) & \\
\midrule
RIFE \cite{rife}             & 2        & 31.48 / 0.124 & 27.84 / 0.173 & 36.29 / 0.039 & 31.84 / 0.073 & 27.18 / 0.135 & 35.15 / 0.037 \\
EMA-VFI \cite{ema-vfi}       & 2        & 32.48 / 0.098 & 28.71 / 0.143 & 36.61 / 0.042 & 32.25 / 0.075 & 27.69 / 0.131 & 35.38 / 0.036 \\
QVI \cite{qvi}               & 4        & 31.29 / 0.122 & 27.51 / 0.181 & 35.88 / 0.044 & 31.61 / 0.080 & 26.93 / 0.134 & 35.60 / 0.033 \\
EQVI \cite{eqvi}             & 4        & 31.70 / 0.119 & 27.88 / 0.175 & 36.11 / 0.042 & 31.85 / 0.077 & 27.15 / 0.132 & 35.71 / 0.031 \\
IQ-VFI \cite{iqvfi}          & 2        & 31.91 / 0.109 & 28.23 / 0.156 & 36.39 / 0.039 & 32.11 / 0.069 & 27.43 / 0.123 & 35.79 / 0.029 \\
GIMM-VFI \cite{gimm}         & 2        & 32.68 / 0.114 & 29.01 / 0.137 & 36.98 / 0.034 & 32.59 / 0.061 & 27.96 / 0.111 & 35.91 / 0.027 \\
CT-VFI \cite{ctvfi}          & 2$^\ast$ & 30.82 / 0.137 & 26.74 / 0.201 & 35.50 / 0.049 & 31.11 / 0.089 & 26.43 / 0.144 & 35.68 / 0.032 \\
CT-VFI (spline) \cite{ctvfi} & $N$      & 32.61 / 0.109 & 28.88 / 0.142 & 36.88 / 0.035 & 32.51 / 0.063 & 27.88 / 0.113 & 36.09 / 0.025 \\
\midrule
Ours-V (linear)              & 2        & 30.90 / 0.133 & 26.78 / 0.197 & 35.63 / 0.047 & 31.20 / 0.087 & 26.51 / 0.141 & 36.21 / 0.023 \\
Ours-V                       & 2        & 32.52 / 0.102 & 28.64 / 0.146 & 36.81 / 0.036 & 32.38 / 0.065 & 27.83 / 0.116 & \textbf{36.24} / \textbf{0.022} \\
Ours-X (linear)              & 2        & 30.94 / 0.132 & 26.84 / 0.194 & 35.68 / 0.046 & 31.25 / 0.086 & 26.60 / 0.138 & 35.83 / 0.029 \\
Ours-X (no LPIPS)            & 2        & 33.15 / 0.091 & 29.71 / 0.125 & 37.31 / 0.031 & 33.05 / 0.056 & 28.52 / 0.101 & 35.91 / 0.029 \\
Ours-X                       & 2        & \textbf{33.21} / \textbf{0.087} & \textbf{29.78} / \textbf{0.119} & \textbf{37.38} / \textbf{0.028} & \textbf{33.11} / \textbf{0.052} & \textbf{28.60} / \textbf{0.095} & 35.95 / 0.027 \\
\bottomrule
\end{tabular*}
\end{table*}

\subsection{Frame quality}

The within-system motion gains largely carry through synthesis (Table~\ref{tab:main}).
Ours-X gains 2.94 dB on X-TEST $16\times$ over its matched linear variant. It is also
0.77 dB above the released GIMM-VFI system, although that external difference should not be
attributed solely to the Hermite parameterization. On SNU-FILM, the matched gap over the
linear variant grows with motion magnitude, from 1.70 dB on Medium to 1.86 on Hard and
2.00 on Extreme, consistent with additional trajectory freedom mattering most away from
near-linear motion. CT-VFI (spline) illustrates the role of the downstream renderer: it
slightly leads GIMM-VFI on the motion metric but trails it on synthesized-frame PSNR.

Dropping the LPIPS term from training costs 0.06 dB and 0.004 LPIPS on X-TEST $8\times$.
That row still leads GIMM-VFI on both metrics on every benchmark except VSF, where it ties on
PSNR and trails on LPIPS, so the perceptual margin is not an artefact of optimising the
metric we report. It also reproduces the widening SNU-FILM trend, at 1.63, 1.80 and 1.92 dB
over the linear variant.

VSF is the other end of that trend and we read it accordingly. Ours-V stays within 0.03 dB
of its linear variant there while gaining 1.32 dB on Extreme, so the added expressiveness
costs nothing where it is unnecessary. The corollary is that our VSF number is carried by
the synthesis stage rather than by the trajectory, which is what a benchmark of near-linear
motion should show.

\subsection{Dependence on trajectory cues}

Endpoint displacement specifies where a pixel starts and ends but is insufficient to
identify its intermediate route. Table~\ref{tab:inputs} probes which inputs the
\emph{trained} coefficient predictor relies on by zeroing channel groups at test time. This
is a dependency analysis, not a retrained causal ablation. 
\begin{table*}[htbp]
\centering
\caption{Runtime (ms) versus number of interpolated frames, $720\times1280$ input on a
single NVIDIA V100. Motion is the trajectory model alone. Total additionally includes flow
reversal, synthesis, and each method's front-end feature extraction, a fixed cost that does
not scale with $K$. Ours uses RAFT and reuses its internal features rather than running a
second extraction pass, which is why our fixed cost is the lowest here.}
\label{tab:cost}
\begin{tabular*}{\textwidth}{@{\extracolsep{\fill}}lcccccc@{}}
\toprule
Method & Params (M) & $K{=}1$ & $K{=}8$ & $K{=}16$ & $K{=}32$ & Slope (ms/frame) \\
\midrule
IQ-VFI \cite{iqvfi}    & 14.2 & 128.3 & 233.9 & 355.2 & 599.5 & 15.2 \\
GIMM-VFI \cite{gimm}   & 12.5 & 126.0 & 218.2 & 324.1 & 535.7 & 13.2 \\
CT-VFI \cite{ctvfi}    & 28.4 & 133.8 & 271.6 & 430.9 & 747.6 & 19.8 \\
\midrule
Ours (motion)          & 0.43 & 2.1   & 2.1   & 2.1   & 2.1   & 0.0  \\
Ours (total)           & 10.8 & 110.5 & 177.5 & 254.2 & 408.4 & 9.6  \\
\bottomrule
\end{tabular*}
\end{table*}

With only flow-derived inputs
retained, the predicted residual magnitude collapses to $0.08$ pixels and EPE returns to the
linear-control level (3.49 versus 3.48). Restoring appearance features increases the learned
residuals and reduces EPE to 1.44; restoring the structure-tensor directional descriptor
further reduces EPE to 0.97. These results show that the jointly trained predictor depends
strongly on appearance-derived information beyond endpoint displacement. Because the
structure tensor is not blur specific, we interpret its contribution as complementary local
anisotropy/orientation rather than direct measurement of physical motion blur.

\begin{table}[htbp]
\centering
\caption{Test-time input-dependence analysis for CoeffHead, Ours-X on X-TEST-Flow
$16\times$. Channel groups are masked in one trained model (no retraining);
$\lVert\delta_i\rVert$ is in post-upsampling pixels.}
\label{tab:inputs}
\begin{tabular}{lcccc}
\toprule
Inputs & PSNR$_f\uparrow$ & EPE$\downarrow$ & $\lVert\delta_0\rVert$ & $\lVert\delta_1\rVert$ \\
\midrule
Flow only                   & 25.04 & 3.49 & 0.08 & 0.08 \\
\;+ appearance $\mathcal S$ & 32.12 & 1.44 & 10.22 & 10.45 \\
\;+ directional $\mathcal B$       & 34.19 & 0.97 & 12.14 & 12.38 \\
\bottomrule
\end{tabular}
\end{table}

We reuse RAFT's convex-upsampling mask ungated in the first row. It redistributes $\delta$
within each $8\times8$ block but cannot create magnitude where the low-resolution
prediction is zero, so the test remains valid.

\subsection{Number of supervised times}

\begin{table}[htbp]
\centering
\caption{Number of supervised query times per frame pair, Ours-X on X-TEST-Flow
$16\times$.}
\label{tab:timesteps}
\begin{tabular}{lcccc}
\toprule
Times per pair & PSNR$_f\uparrow$ & EPE$\downarrow$ & $\lVert\delta_i\rVert$ & $\rho(\delta_0,\delta_1)$ \\
\midrule
1 ($t=1/2$) & 25.07 & 3.48 & 12.14 & 0.98 \\
3           & 31.47 & 1.66 & 11.78 & 0.58 \\
5           & 34.19 & 0.97 & 12.14 & 0.18 \\
\bottomrule
\end{tabular}
\end{table}

Table~\ref{tab:timesteps} varies the number of supervised times per pair at fixed gradient
steps, so the comparison changes supervision structure rather than optimization budget.
With only the midpoint, the residuals are almost perfectly correlated ($\rho=0.98$) while
EPE remains at the linear-control level. This is consistent with
Propositions~2--3: at $t=1/2$ the common component $\bar\delta$ in
Eq.~\eqref{eq:bound} vanishes exactly, so large nearly shared endpoint residuals can be
weakly constrained by midpoint reconstruction, and one time cannot uniquely identify both
trajectory coefficients. With five supervised times, the correlation falls to 0.18 and EPE
drops to 0.97. We interpret this trend as empirical evidence consistent with the
representation-level identifiability analysis, not as proof that reconstruction removes all
physical two-frame ambiguity.

This bears on a prior negative result. IQ-VFI \cite{iqvfi} reports no gain from an
acceleration branch trained on reconstruction loss alone, on Vimeo triplets. Their model
has one free coefficient per pixel and a single $t$ determines it, so the rank deficiency
of Sec.~\ref{sec:trajectory} does not apply directly. What the row above shows is a milder
version of the same problem: at one timestep nothing forces a predicted correction to hold
at any other time, and our own residuals stay large while remaining unusable. We read the
IQ-VFI result as a consequence of the supervision structure rather than of photometric
supervision being too weak, but the two settings are not interchangeable and we do not
claim to have reproduced theirs.

\subsection{Cost}

Table~\ref{tab:cost} shows our trajectory-model cost flat at 2.1 ms across all $K$: the
query-independent coefficient head runs once per frame pair, and each additional trajectory
time costs three tensor products. The
comparison methods grow at 15.2, 13.2, and 19.8 ms per interpolated frame. Those slopes are
for the full pipeline rather than for the motion model alone, since reversal and synthesis
also run per $t$ and the two cannot be separated from an end-to-end timing. The like-for-like
statement is that our own pipeline grows at 9.6 ms per frame while the trajectory contributes
nothing to that figure, and that end to end at $K = 32$ we are $1.3\times$ faster than
GIMM-VFI on the same input.

The amortisation therefore applies to the trajectory model alone. Reversal and synthesis
still run per $t$ for every method, ours included, which is where the remaining per-frame
cost lives.


\section{Conclusion}
\label{sec:conclusion}

HermiteVFI treats two-frame intermediate motion as one explicit cubic trajectory rather than
a collection of independent query-time predictions. Endpoint displacement fixes the two
trajectory positions, while two latent endpoint derivatives determine the remaining degrees
of freedom. A cubic is the minimum polynomial order that can independently satisfy those
four boundary quantities. In residual form, the measured linear path is an exact nested
special case and the correction vanishes at both displacement endpoints for any network
output. The learned coefficients are independent of $t$, so one coefficient prediction fixes
the whole motion curve and later trajectory evaluations are closed-form polynomial
operations. Our identifiability analysis further explains why a single interior supervision
time is insufficient for the two free coefficients and why multi-time supervision is a
natural consequence of the representation.

\paragraph{Limitations and future work.}
Two RGB endpoints do not uniquely determine the physical path between them. HermiteVFI does
not remove this fundamental ambiguity; it restricts the solution to a low-dimensional,
endpoint-consistent cubic family and learns an image-conditioned prior over its latent
endpoint derivatives. Accordingly, these ``velocities'' are reconstruction-optimal
trajectory parameters rather than direct measurements of physical instantaneous velocity.
The endpoint guarantee applies to the displacement trajectory, not exact RGB reconstruction.
Only trajectory-coefficient prediction is fully amortized; reversal, refinement, and
synthesis still run at every $t$. A cubic also has limited temporal complexity, and
appearance-derived directional cues may be weak in globally sharp or textureless regions.
Higher-order endpoint-vanishing bases and amortized synthesis are promising future work.

{
    \small
    \bibliographystyle{ieeenat_fullname}
    \bibliography{main}
}

\clearpage






\setcounter{page}{1}
\maketitlesupplementary
\appendix

\section{Method Details}

\subsection{Derivation of the residual form}
\label{sup:derivation}

The cubic Hermite basis on $[0,1]$ is
\begin{align}
h_{00}(t) &= 2t^3 - 3t^2 + 1, & h_{10}(t) &= t^3 - 2t^2 + t, \nonumber\\
h_{01}(t) &= -2t^3 + 3t^2,    & h_{11}(t) &= t^3 - t^2 .
\end{align}
With endpoint positions $\Phi(0)=\mathbf{0}$, $\Phi(1)=F$ and endpoint tangents $m_0,
m_1$, the interpolant is $\Phi(t) = h_{00}\Phi(0) + h_{10}m_0 + h_{01}\Phi(1) +
h_{11}m_1$; the $h_{00}$ term drops, giving Eq.~(1) of the main paper. Substituting
$m_i = F + \delta_i$ and grouping,
\begin{equation}
\Phi(t) = \underbrace{\big(h_{10} + h_{01} + h_{11}\big)}_{=\,t} F
+ h_{10}\,\delta_0 + h_{11}\,\delta_1 .
\end{equation}
The bracket collapses because the cubic and quadratic terms cancel,
$(1\!-\!2\!+\!1)t^3 + (-2\!+\!3\!-\!1)t^2 + t = t$. The two residual coefficients
re-express in the basis $\beta_2(s) = s^2\!-\!s$, $\beta_3(s) = s^3\!-\!s$ as
\begin{equation}
h_{10} = \beta_3 - 2\beta_2, \qquad h_{11} = \beta_3 - \beta_2 ,
\end{equation}
so that $\Phi(t) = tF + \beta_2(-2\delta_0 - \delta_1) + \beta_3(\delta_0 + \delta_1)$,
which is Eq.~(2) with the coefficients of Eq.~(3). Inverting the coefficient map,
$\delta_0 = -A - B$ and $\delta_1 = A + 2B$.

\subsection{Minimum degree, identifiability, and deviation bound}
\label{sup:theory}

\paragraph{Minimum degree.}
For each motion component, the trajectory is constrained by two endpoint positions and two
endpoint derivatives: $\Phi(0)$, $\Phi(1)$, $\dot\Phi(0)$, and $\dot\Phi(1)$. A degree-$d$
polynomial has $d+1$ scalar coefficients, so four independent boundary quantities require
$d\geq3$. The cubic Hermite form is therefore the minimum polynomial degree that supports
these four constraints without tying the two endpoint derivatives together.

\paragraph{Two-time identifiability.}
The residual trajectory is
$y(t)=\Phi(t)-tF=\beta_2(t)A+\beta_3(t)B$. Because
$\beta_3(t)=\beta_2(t)(t+1)$, two interior observations satisfy
\begin{equation}
\begin{bmatrix}y(t_1)\\y(t_2)\end{bmatrix}
=
\begin{bmatrix}
\beta_2(t_1) & \beta_3(t_1)\\
\beta_2(t_2) & \beta_3(t_2)
\end{bmatrix}
\begin{bmatrix}A\\B\end{bmatrix}.
\label{sup:eq:ident}
\end{equation}
The determinant of the coefficient matrix is
\begin{equation}
\det = \beta_2(t_1)\beta_2(t_2)(t_2-t_1).
\end{equation}
For $t_1,t_2\in(0,1)$, $\beta_2(t_i)=t_i(t_i-1)\neq0$; hence any distinct
$t_1\neq t_2$ give a nonzero determinant and uniquely determine $A,B$ at the
representation level. One interior time supplies only one equation and leaves a one-
dimensional solution set. This algebraic statement assumes direct trajectory observation;
image reconstruction additionally contains visibility, correspondence, occlusion, and
synthesis ambiguity.

\paragraph{Corrected deviation extrema.}
With $\bar\delta=(\delta_0+\delta_1)/2$ and
$\Delta=(\delta_1-\delta_0)/2$,
\begin{equation}
\Phi(t)-tF=\bar\delta\,t(2t-1)(t-1)+\Delta(t^2-t).
\end{equation}
For $g_c(t)=t(2t-1)(t-1)$, $g_c'(t)=6t^2-6t+1$, whose interior extrema occur at
$t=(3\pm\sqrt3)/6$ and have magnitude $\sqrt3/18\approx0.0962$. For
$g_d(t)=t^2-t$, the unique interior extremum is $t=1/2$ with magnitude $1/4$.
Consequently,
\begin{equation}
\sup_{t\in[0,1]}\|\Phi(t)-tF\|
\leq \frac{\sqrt3}{18}\|\bar\delta\|+\frac14\|\Delta\|,
\end{equation}
for any norm satisfying the triangle inequality. The differential basis therefore has
approximately $2.6\times$ greater maximum amplitude than the common basis.

\subsection{Derivation of the velocity consistency constraint}
\label{sup:velderiv}

Write $s = 1-t$ in the backward trajectory. Since $\beta_2'(0) = \beta_3'(0) = -1$ and
$\mathrm{d}s/\mathrm{d}t = -1$,
\begin{equation}
\dot\Phi'(1) = -\big[F' + \beta_2'(0)A' + \beta_3'(0)B'\big]
= -\big[F' - A' - B'\big] = -m'_0 ,
\end{equation}
using $F' - A' - B' = F' + \delta'_0 = m'_0$ from Sec.~\ref{sup:derivation}. A pixel at
$\mathbf{x}$ on the grid of $I_0$ occupies $\mathbf{x} + F(\mathbf{x})$ on the grid of
$I_1$, so the velocity the backward model reports there, transported back by
$\overleftarrow{\omega}(\cdot, F)$, must equal $-m_1(\mathbf{x})$. The main-paper
velocity-consistency loss penalizes this discrepancy, weighted by the same reliability map
$\alpha$ used by the occlusion gate, which is small where
the forward--backward check disagrees. Occluded pixels, where no correspondence exists,
therefore contribute little. Each term takes the map belonging to the lattice it is
evaluated on: the first is a lattice-0 quantity and takes $\alpha$, the second is a
lattice-1 quantity and takes $\alpha'$.

\subsection{Directional appearance descriptor}
\label{sup:blur}

Let $J_i = \mathcal{G}_{5\times5} * (\nabla I_i \nabla I_i^{\!\top})$, computed on the
greyscale image with Sobel gradients and Gaussian width $\sigma_J = 1.0$, with entries
$J_{xx}, J_{xy}, J_{yy}$. Writing $d = J_{xx} - J_{yy}$ and $r = \sqrt{d^2 +
4J_{xy}^{2}}$, the eigenvalues are
\begin{equation}
\lambda_{1,2} = \tfrac12\big(J_{xx} + J_{yy} \pm r\big),
\end{equation}
and the double-angle orientation follows without an $\arctan$,
\begin{equation}
\cos 2\theta = \frac{d}{r + \epsilon}, \qquad
\sin 2\theta = \frac{2J_{xy}}{r + \epsilon} .
\end{equation}
The double-angle encoding is invariant to the $\pi$ ambiguity of the dominant
orientation; direction sign is not recoverable from $J_i$ alone and is left to the velocity
head, which also receives $F$. We use $\epsilon = 10^{-6}$ and clamp the anisotropy ratio
$\lambda_1/(\lambda_2+\epsilon)$ to $[1,50]$ before normalization. This descriptor is
not specific to motion blur: it measures local anisotropy/orientation and can also respond
to ordinary oriented image structure.

\subsection{Convex upsampling}
\label{sup:upsample}

Upsampling is applied to the already-gated low-resolution residual. RAFT's mask
$W_0 \in \mathbb{R}^{\frac{H}{8}\times\frac{W}{8}\times 576}$ is reshaped to
$(9, 8, 8, \frac{H}{8}, \frac{W}{8})$ and normalized by softmax over the first axis. Each
full-resolution output at sub-position $(p,q) \in \{0,\dots,7\}^2$ within coarse cell
$\mathbf{v}$ is a convex combination of the $3\times3$ coarse neighbourhood,
\begin{equation}
\delta_i\big(8\mathbf{v} + (p,q)\big) = 8 \sum_{k=1}^{9}
\mathrm{softmax}(W_0)_{k,p,q,\mathbf{v}} \;\,
\delta_i^{\downarrow 8}(\mathbf{v} + \mathbf{n}_k),
\end{equation}
with $\mathbf{n}_k$ enumerating the neighbourhood offsets. The factor $8$ converts from
$1/8$-resolution pixel units to full-resolution pixels. Because the weights are shared
with $F$, the upsampled residual cannot spread across a motion boundary that $F$
respects.

\begin{algorithm*}[h]
\caption{HermiteVFI inference}
\label{alg:infer}
\begin{algorithmic}[1]
\Require frames $I_0, I_1$; query times $\{t_k\}_{k=1}^{K}$; frozen flow estimator $\mathcal{E}$
\Ensure interpolated frames $\{\hat I_{t_k}\}$
\State $F, c_0, h_0, W_0 \gets \mathcal{E}(I_0, I_1)$ \Comment{no gradient}
\State $F', c_1, h_1, W_1 \gets \mathcal{E}(I_1, I_0)$
\State $U \gets \|F + \overleftarrow{\omega}(F',F)\|_1$, \quad
       $Z \gets -\|I_0 - \overleftarrow{\omega}(I_1,F)\|_1$ \Comment{$U', Z'$ likewise}
\State $\mathcal{B}_0, \mathcal{B}_1 \gets$ structure-tensor directional descriptors
\For{$j \in \{0,1\}$} \Comment{shared weights, both grids}
  \State $\mathcal{S}_j \gets \mathrm{AppNet}\big([\,I_j,\; \overleftarrow{\omega}(I_{1-j},F_j),\; \mathcal{B}_j,\; \overleftarrow{\omega}(\mathcal{B}_{1-j},F_j)\,]\big)$
  \State $\Psi_j \gets [\, c_j,\; h_j,\; \tfrac18 F_j^{\downarrow 8},\; \tfrac18 \overleftarrow{\omega}(F_{1-j},F_j)^{\downarrow 8},\; U_j^{\downarrow 8},\; \mathcal{S}_j \,]$
  \State $\delta_0^{\downarrow 8}, \delta_1^{\downarrow 8} \gets \mathcal{C}(\Psi_j)$, \quad
         $\delta_i^{\downarrow 8} \gets \mathrm{sig}(w_1 - w_2 U_j^{\downarrow 8})\,\delta_i^{\downarrow 8}$
  \State $\delta_i \gets \mathrm{ConvexUp}\big(8\delta_i^{\downarrow 8},\, \mathrm{softmax}(W_j)\big)$, \quad
         $A_j \gets -2\delta_0 - \delta_1$, \quad $B_j \gets \delta_0 + \delta_1$
\EndFor
\For{$k = 1$ \textbf{to} $K$}
  \State $\Phi \gets t_k F + \beta_2(t_k)A + \beta_3(t_k)B$, \quad
         $\Phi' \gets (1\!-\!t_k)F' + \beta_2(1\!-\!t_k)A' + \beta_3(1\!-\!t_k)B'$
  \State $\tilde G_0, \Omega_0 \gets \overrightarrow{\omega}(-\Phi, \Phi, Z)$, \quad
         $\tilde G_1, \Omega_1 \gets \overrightarrow{\omega}(-\Phi', \Phi', Z')$
  \State $G_i \gets \tilde G_i + \mathrm{RefineNet}(\tilde G_0, \tilde G_1, \Omega_0, \Omega_1)$
  \State $I_{t\to i} \gets \overleftarrow{\omega}(I_i, G_i)$, \quad
         $M, R \gets \mathrm{SynthNet}(I_{t\to0}, I_{t\to1}, G_0, G_1)$
  \State $\hat I_{t_k} \gets M \odot I_{t\to0} + (1\!-\!M)\odot I_{t\to1} + R$
\EndFor
\end{algorithmic}
\end{algorithm*}

\subsection{Splatting and warping}
\label{sup:splat}

The target-grid transfer uses softmax splatting~\cite{softsplat}. For a target pixel $\mathbf{u}$, let
$\mathcal{N}(\mathbf{u})$ be its four-pixel bilinear neighbourhood, $w(\mathbf{x}) \in
[0,1]$ the bilinear weight of a splatted source pixel, and
\begin{equation}
\mathcal{P}(\mathbf{u}) = \big\{\, \mathbf{x} \;:\;
\mathbf{x} + \Phi(t)(\mathbf{x}) \in \mathcal{N}(\mathbf{u}) \,\big\}
\end{equation}
the set of source pixels landing on it. The transferred field and its accumulated weight
are
\begin{equation}
\tilde G_0(\mathbf{u}) =
\frac{\sum_{\mathbf{x} \in \mathcal{P}(\mathbf{u})}
w(\mathbf{x})\,e^{Z(\mathbf{x})}\,\big(\!-\!\Phi(t)(\mathbf{x})\big)}
{\sum_{\mathbf{x} \in \mathcal{P}(\mathbf{u})} w(\mathbf{x})\,e^{Z(\mathbf{x})}} ,
\end{equation}
\begin{equation}
\Omega_0(\mathbf{u}) = \sum_{\mathbf{x} \in \mathcal{P}(\mathbf{u})}
w(\mathbf{x})\,e^{Z(\mathbf{x})} .
\end{equation}
The negated numerator points the transferred field from $t$ back to $0$. Where
$\Omega_0(\mathbf{u}) = 0$ nothing landed on $\mathbf{u}$ and the quotient is set to zero;
these are the holes filled by RefineNet in the main pipeline. $\tilde G_1$ and $\Omega_1$ follow from $\Phi'(t)$ and
$Z'$. Backward warping is bilinear resampling,
\begin{equation}
I_{t\to i}(x,y) = I_i\big(x + G_i^{h}(x,y),\; y + G_i^{v}(x,y)\big),
\end{equation}
with $G_i^{h}, G_i^{v}$ the horizontal and vertical components of $G_i$.

\subsection{Photometric bundle}
\label{sup:photo}

The photometric bundle used by the reconstruction objective is
\begin{equation}
\mathcal{L}_{\text{photo}}(\hat I, I) = \rho(\hat I - I)
+ \mathcal{L}_{\text{lap}}(\hat I, I)
+ \mathcal{L}_{\text{cen}}(\hat I, I)
+ \mathcal{L}_{\text{lpips}}(\hat I, I),
\end{equation}
with all four terms at equal weight. $\rho(x) = \sqrt{x^2 + \varepsilon^2}$ is the
Charbonnier penalty with $\varepsilon = 10^{-3}$; $\mathcal{L}_{\text{lap}}$ is an
$\ell_1$ loss over a five-level Laplacian pyramid; $\mathcal{L}_{\text{cen}}$ is the
census (ternary) loss on $7\times7$ patches~\cite{rife}; and
$\mathcal{L}_{\text{lpips}}$ uses an AlexNet backbone~\cite{lpips}. The motion scale used by the velocity-consistency objective is
$s = \max(\|F\|_\infty, \|F'\|_\infty)$, matching the main paper.
\section{Network Architectures}
\label{sup:arch}

All velocity-stage modules operate at $1/8$ resolution. Convolutions use PReLU activation
unless noted, and a residual unit is two $3\times3$ convolutions with a skip connection.
Table~\ref{tab:arch} lists the modules.

\begin{table}[h]
\centering
\scriptsize
\setlength{\tabcolsep}{2pt}
\begin{tabular}{@{}llcr@{}}
\toprule
Module & Structure & Channels & Params \\
\midrule
AppNet          & $3\times$ stride-2 conv     & $12\!\to\!32\!\to\!64\!\to\!64$ & 60K \\
Fusion          & $1\times1$ conv             & $325 \to 96$ & 31K \\
$\mathcal{C}$ body   & $2\times$ residual unit & $96 \to 96$  & 332K \\
$\mathcal{C}$ output & $3\times3$ conv, zero-init & $96 \to 4$ & 3.5K \\
Occlusion gate  & two scalars                 & $1 \to 1$    & 2 \\
\midrule
\multicolumn{3}{@{}l}{Total, velocity stage}  & \textbf{427K} \\
\bottomrule
\end{tabular}
\caption{Velocity estimation modules. The upsampling mask is reused from RAFT and
contributes no parameters. The stage is shared between the forward and backward passes
and evaluated once per frame pair.}
\label{tab:arch}
\end{table}

AppNet takes $12$ channels: $I_0$, $\overleftarrow{\omega}(I_1,F)$, and the two directional
descriptors ($3\!+\!3\!+\!3\!+\!3$). The fusion input is $325$ channels: $c_0$ and $h_0$
at $128$ each, two flow fields at $2$ each, $U$ at $1$, and $\mathcal{S}_0$ at $64$.

\paragraph{Synthesis stages.}
The two stages after the trajectory are our own and follow splatting-based
synthesis~\cite{softsplat}. RefineNet takes $6$ channels ($\tilde G_0, \tilde G_1,
\Omega_0, \Omega_1$) and returns the $4$-channel flow correction; SynthNet
takes $10$ channels ($I_{t\to0}, I_{t\to1}, G_0, G_1$) and returns $M$ and $R$ as
$1\!+\!3$. Unlike the velocity stage these run at full resolution and once per query time,
and together they account for the balance of the $10.8$M parameters reported in Table~5
against the $427$K of Table~\ref{tab:arch}.

\begin{algorithm*}[h]
\caption{Training iteration, matching the main-paper objective}
\label{alg:train}
\begin{algorithmic}[1]
\Require clip with endpoints $I_0,I_1$ and interior frames $\{I_t\}_{t\in\mathcal{T}}$
\State run lines 1--10 of Alg.~\ref{alg:infer} \Comment{gated residuals predicted once per clip}
\State $m_i\gets F+\delta_i$, \quad $m'_i\gets F'+\delta'_i$
\State evaluate full-resolution reliability maps $\alpha,\alpha'$ with the same gate function
\State $s\gets\max(\|F\|_\infty,\|F'\|_\infty)$
\State $\mathcal{L}_{\mathrm{vel}}\gets
\tfrac12\langle\alpha\odot|m_1+\overleftarrow{\omega}(m'_0,F)|/s\rangle
+\tfrac12\langle\alpha'\odot|m_0+\overleftarrow{\omega}(m'_1,F')|/s\rangle$
\State $\mathcal{L}_{\mathrm{rec}}\gets0$
\For{$t\in\mathcal{T}$}
  \State obtain $I_{t\to0},I_{t\to1},M,R,\hat I_t$ from lines 12--16 of Alg.~\ref{alg:infer}
  \State $\tilde I_t\gets M\odot I_{t\to0}+(1-M)\odot I_{t\to1}$
  \State $\mathcal{L}_{\mathrm{rec}}\mathrel{{+}{=}}
  \big[\mathcal{L}_{\mathrm{photo}}(\hat I_t,I_t)+
  \tfrac12\mathcal{L}_{\mathrm{photo}}(\tilde I_t,I_t)\big]/|\mathcal{T}|$
\EndFor
\State backpropagate $\mathcal{L}_{\mathrm{rec}}+\lambda_{\mathrm{vel}}\mathcal{L}_{\mathrm{vel}}$
       \Comment{flow estimator frozen}
\end{algorithmic}
\end{algorithm*}

\section{Algorithms}
\label{sup:algo}

Algorithm~\ref{alg:infer} gives inference for $K$ query times. Lines 1--10 execute once
per frame pair; only the second loop scales with $K$, and within it only line~12 concerns
motion. Algorithm~\ref{alg:train} gives one training iteration.

\section{Input-Dependence Gates}
\label{sup:gates}

The input-dependence analysis in the main paper is evaluated on a single trained checkpoint
by zeroing channel groups of $\Psi_j$ at test time, without retraining
(Table~\ref{tab:gates}). It therefore measures dependence of the jointly trained predictor,
not the causal effect of independently trained input subsets.

\begin{table}[ht]
\centering
\small
\setlength{\tabcolsep}{4pt}
\begin{tabular}{@{}lll@{}}
\toprule
Gate & Zeroed & Retained \\
\midrule
Full          & ---                    & all \\
No appearance & $\mathcal{S}_i$        & $c_i, h_i, F, F', U$ \\
Flow only     & $\mathcal{S}_i, c_i, h_i$ & $F, \overleftarrow{\omega}(F',F), U$ \\
No directional $\mathcal{B}$ & $\mathcal{B}_i$ channels   & all else \\
\bottomrule
\end{tabular}
\caption{Input-dependence gates. Each row masks a channel group of $\Psi_j$ at test time in the same trained checkpoint.}
\label{tab:gates}
\end{table}

The hidden state $h_i$ is initialized from RAFT's context encoder and is therefore
appearance-contaminated; it is zeroed in the flow-only gate for the ablation to be
interpretable. One residual dependency remains: the upsampling mask $W_j$ also derives
from that encoder and is not gated. It redistributes $\delta$ within an $8\times8$ block
and cannot create magnitude where $\delta^{\downarrow 8}=0$, so the flow-only gate still
reveals whether the trained residual output survives when only flow-derived inputs are
retained. We report $\|\delta_0\|_1$ and $\|\delta_1\|_1$ alongside the motion
metrics for every gate.

\section{Qualitative Results}
\begin{figure}[htbp]
    \centering
    \includegraphics[width=1\linewidth]{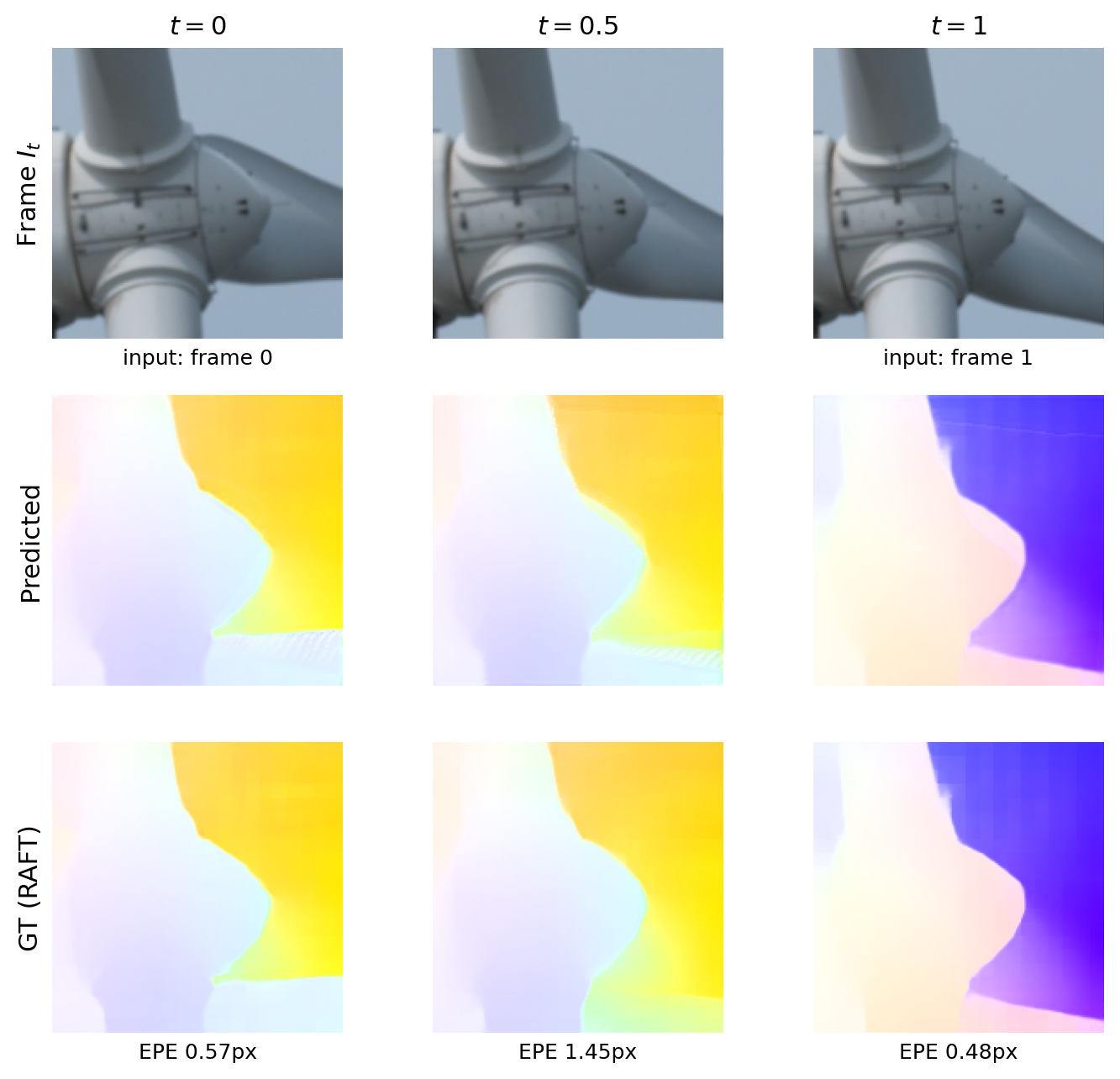}
    \label{fig:qualitative}
\end{figure}
\begin{figure}[htbp]
    \centering
    \includegraphics[width=1\linewidth]{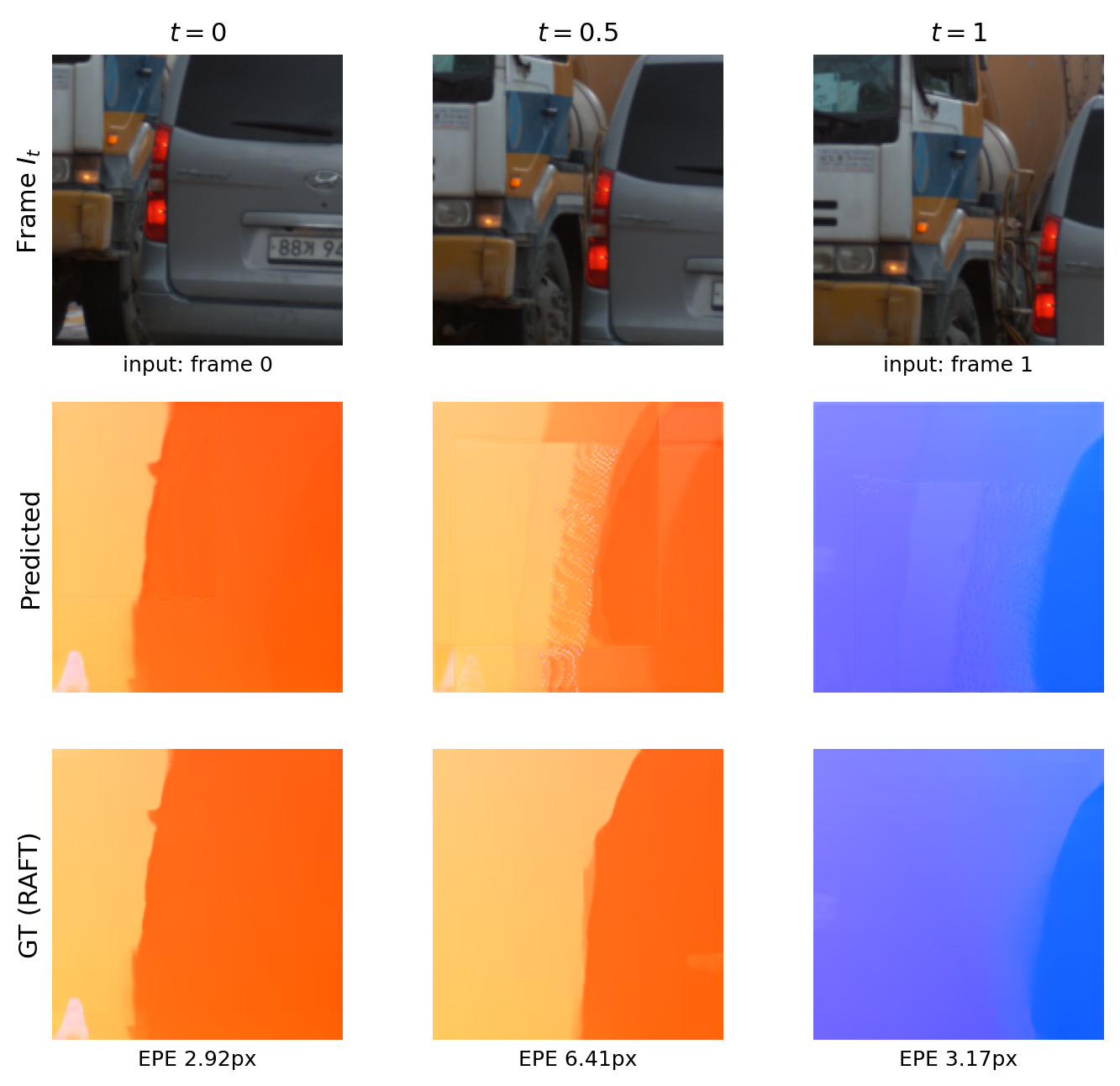}
    \label{fig:qualitative2}
\end{figure}
\begin{figure}[htbp]
    \centering
    \includegraphics[width=1\linewidth]{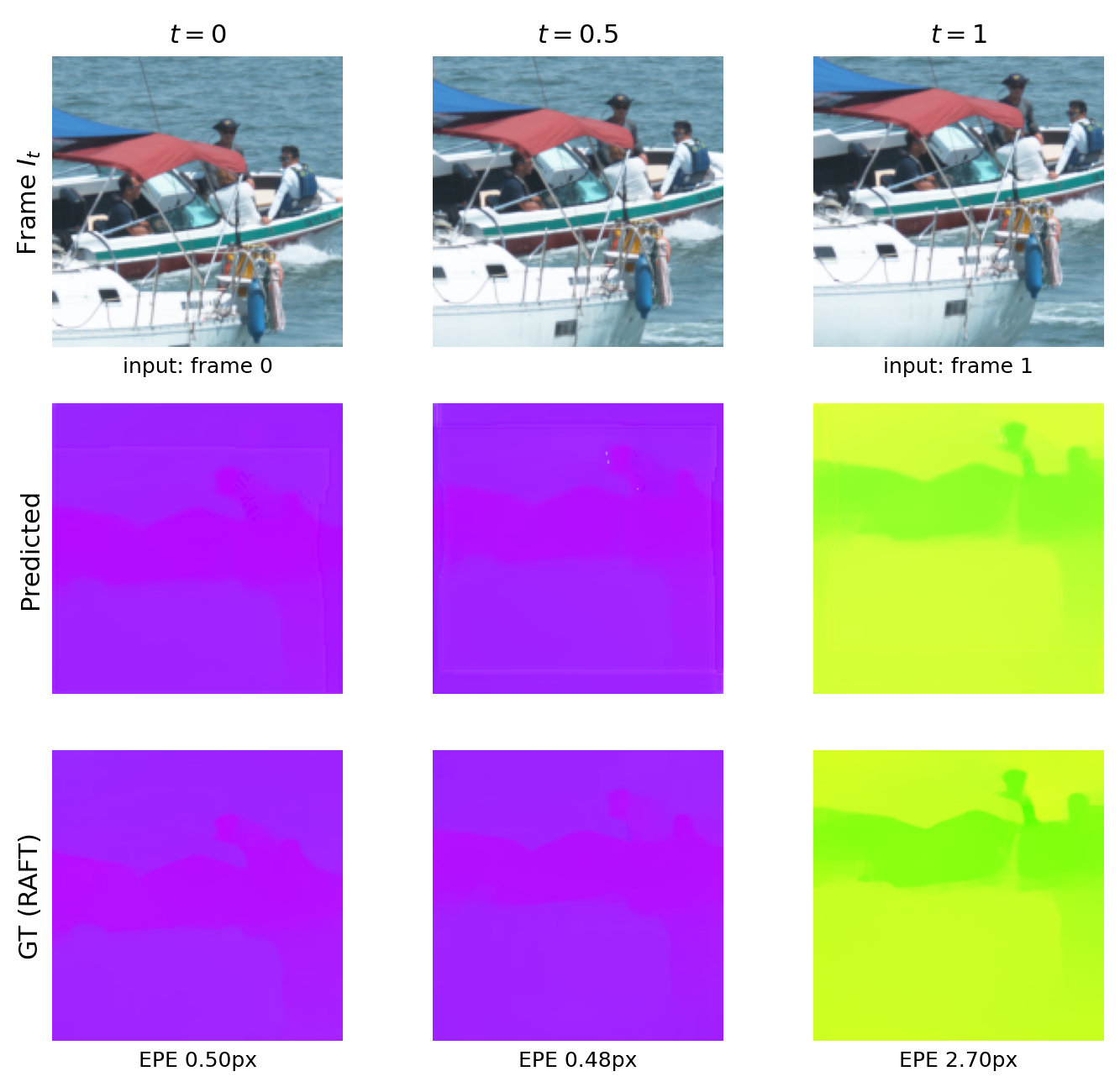}
    \label{fig:qualitative3}
\end{figure}
\begin{figure}[htbp]
    \centering
    \includegraphics[width=1\linewidth]{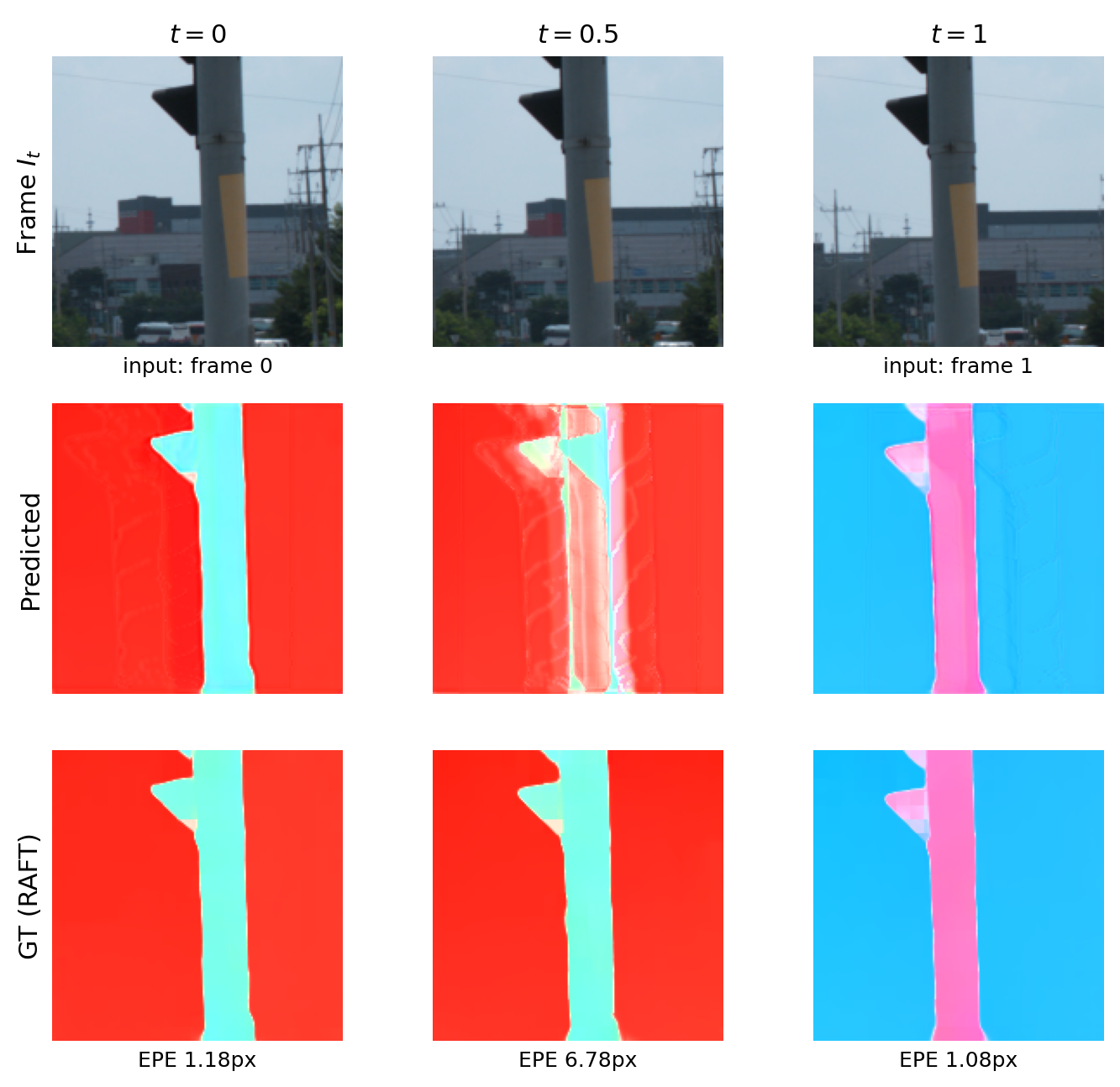}
    \label{fig:qualitative4}
\end{figure}
\begin{figure}[htbp]
    \centering
    \includegraphics[width=1\linewidth]{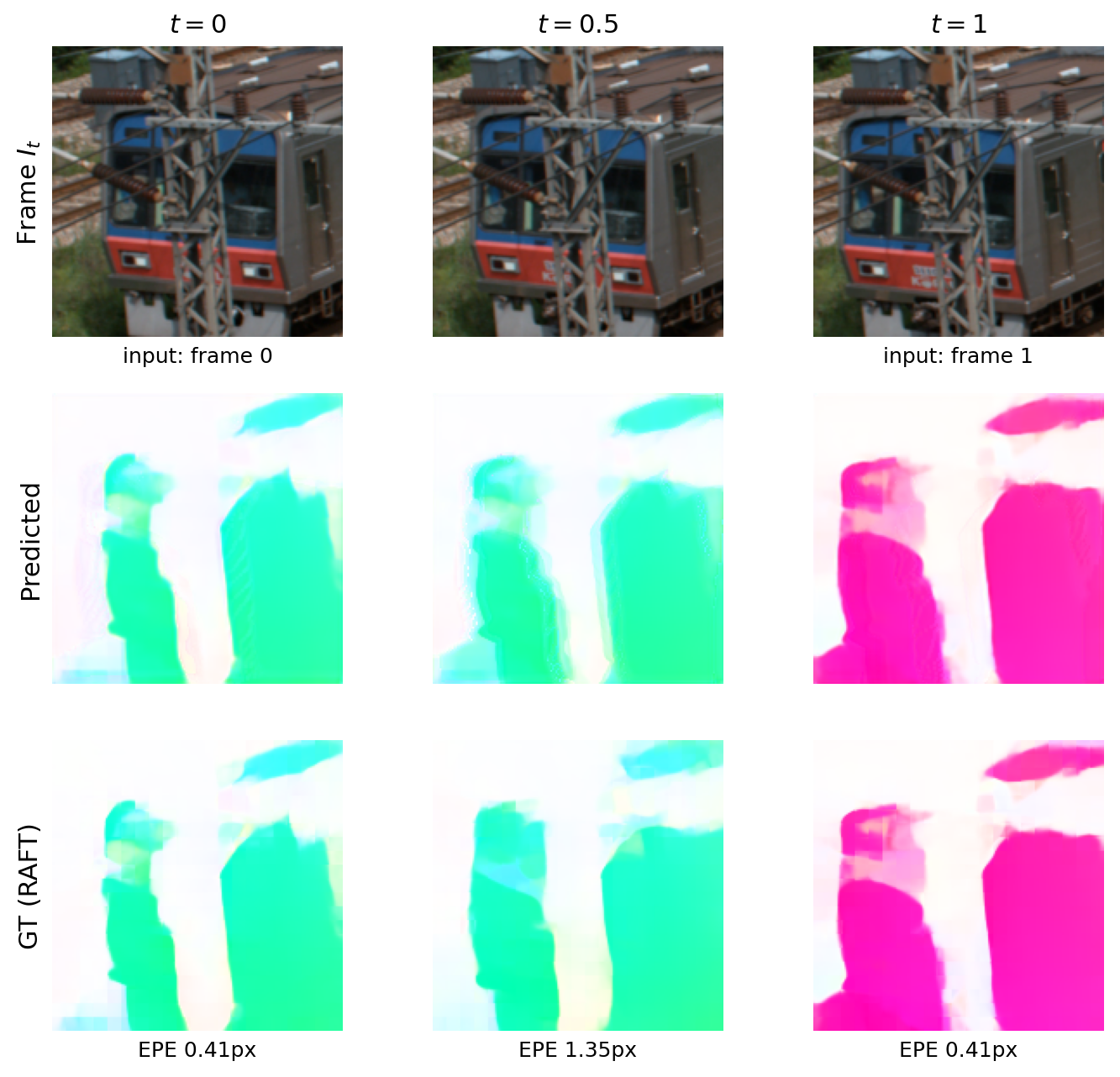}
    \label{fig:qualitative5}
\end{figure}




\end{document}